\documentclass[final,OA]{AAAI-Std}

\def\artid{0001}%

\usepackage{times}
\usepackage{soul,xcolor}
\usepackage{url}
\usepackage[utf8]{inputenc}

\usepackage{graphicx}
\usepackage{amssymb}
\usepackage{amsmath}
\usepackage{amsthm}
\usepackage{booktabs}
\usepackage{algorithm}
\usepackage{refcount}
\usepackage{changepage}
\usepackage[switch]{lineno}
\usepackage{bbding}
\usepackage{graphicx}

\usepackage{mathrsfs}

\usepackage{tabularx}
\usepackage{lineno}
\usepackage{epsfig}
\usepackage{float}
\usepackage{array}
\usepackage{diagbox}
\usepackage{epstopdf}
\usepackage{amsfonts} 
\usepackage{algpseudocode}
\usepackage{threeparttable}
\usepackage{dsfont}
\usepackage{subfigure}
\usepackage{multirow}
\usepackage{longtable}
\usepackage{booktabs}
\usepackage{amsthm}
\usepackage{footnote}
\usepackage{ifpdf}

\begin{document}

\title{Foundation Models for Anomaly Detection: Vision and Challenges}

\author{Jing Ren}
\author{Tao Tang}
\author{Hong Jia}
\author{Ziqi Xu}
\author{Haytham Fayek}
\author{Xiaodong Li}
\author{Suyu Ma}
\author{Xiwei Xu}
\author{Feng Xia}

\address{Jing Ren, Ziqi Xu, Haytham Fayek, Xiaodong Li, and Feng Xia are with the School of Computing Technologies, RMIT University, Melbourne, VIC 3000, Australia (jing.ren@ieee.org; ziqi.xu@rmit.edu.au; haytham.fayek@ieee.org; xiaodong.li@rmit.edu.au; f.xia@ieee.org). Tao Tang is with STEM, the University of South Australia, Adelaide, South Australia 5001, Australia (tao.tang@ieee.org). Hong Jia is with the School of Computing and Information Systems, University of Melbourne, Melbourne, VIC 3010, Australia (hong.jia@unimelb.edu.au). Suyu Ma and Xiwei Xu are with CSIRO's Data61, Australia (suyu.ma@data61.csiro.au, xiwei.xu@data61.csiro.au).}

\abstract[Abstract]{As data continues to grow in volume and complexity across domains such as finance, manufacturing, and healthcare, effective anomaly detection is essential for identifying irregular patterns that may signal critical issues. Recently, foundation models (FMs) have emerged as a powerful tool for advancing anomaly detection. They have demonstrated unprecedented capabilities in enhancing anomaly identification, generating detailed data descriptions, and providing visual explanations. This survey presents the first comprehensive review of recent advancements in FM-based anomaly detection. We propose a novel taxonomy that classifies FMs into three categories based on their roles in anomaly detection tasks, i.e., as encoders, detectors, or interpreters. We provide a systematic analysis of state-of-the-art methods and discuss key challenges in leveraging FMs for improved anomaly detection. We also outline future research directions in this rapidly evolving field.}

\maketitle
	\section{Introduction}
	
	Anomaly detection, also known as outlier detection, is the process of identifying patterns or events in data that significantly deviate from expected behavior~\citep{RenGL4AA2023}. 
	This process is vital across various fields, including, e.g., finance~\citep{park2024enhancing}, business~\citep{guan2025dabl}, manufacturing~\citep{gu2024anomalygpt}, social media~\citep{duan2025llm} and healthcare~\citep{moor2023foundation}. Traditionally, classical methods like $k$-nearest neighbors and clustering algorithms have been employed for this purpose. However, recent advancements in machine learning, such as deep learning, one-class classification, self-supervised learning, and generative adversarial networks, have revolutionized this field. These modern techniques have proved extremely successful in detecting or even predicting anomalies from normal data, making them the primary approaches of anomaly analysis in a wide range of applications~\citep{RenDVAD2021,LiuDGL4ACD2022}.
	
	
	Foundation models (FMs) are a class of large-scale, pre-trained machine learning models that are trained on massive, diverse datasets using self-supervised or unsupervised learning~\citep{bommasani2021opportunities}. In recent years, they have shown impressive performance in learning a wide range of representations and tasks, such as translation, summarization, and question answering, with notable examples including include GPT-4, Gemini and CLIP. 
	Trained on extensive and diverse datasets, FMs comprise a large number of parameters that enable them to capture intricate patterns and subtle nuances within data. This ability to model such complexity makes them particularly well-suited for anomaly detection, where identifying deviations from normal behavior is essential.
	
	
	FMs offer several advantages for anomaly detection, driven by advancements in computational hardware, model architectures, and access to large-scale training data. 
	A key advantage is their capacity for in-context learning, which enables seamless adaptation to entirely new anomaly detection tasks using only natural language instructions~\citep{bakumenko2024advancing}. 
	This eliminates the need for extensive task-specific training for real-world applications, enabling a more flexible and efficient approach. Additionally, many modern FMs integrate multiple data modalities~\citep{lu2022unified}, such as text, images, and time-series data, which is crucial for detecting anomalies in complex scenarios involving diverse data types. 
	For instance, multimodal models such as Gato~\citep{DBLP:journals/tmlr/ReedZPCNBGSKSEBREHCHVBF22}, can perform tasks ranging from image captioning to robotic control, exemplifying the potential of such models as generalist agents. 
	Furthermore, the application of FMs in anomaly detection can enhance explainability, a critical requirement in high-stakes domains such as healthcare, finance, and cybersecurity, where understanding the rationale behind detected anomalies is critical.

	FMs are increasingly being applied to anomaly detection tasks. However, systematic reviews that thoroughly examine and depict this field are still limited. While some works have explored the use of large language models (LLMs) for anomaly detection and have made initial attempts to formalize the research landscape, few provide a comprehensive summary of current progress or address the key challenges in this area. For instance, Su et al.~\citep{su2024large} review the application of LLMs in forecasting and anomaly detection but do not provide a taxonomy specifically tailored to show how FMs are applied to anomaly detection tasks. Similarly, Xu et al.~\citep{xu2024large} focus on anomaly and out-of-distribution detection with LLMs but overlooks other FMs capable of processing diverse data modalities. This highlights the need for a more systematic and inclusive review. In this survey, we focus on methodologies that leverage FMs for anomaly detection to provide a clearer and more structured perspective by categorizing the models based on the roles FMs play in the detection process—specifically as encoders, detectors, or interpreters. This framework offers a comprehensive overview of the field, presents key insights and advancements in the field, and identifies potential directions for future research. 
	

	Our main contributions include:
	\begin{itemize}
		\item \textbf{A novel taxonomy.} We propose a structured taxonomy that categorizes the role of FMs in anomaly detection into three primary functions: encoder, detector, and interpreter. This classification provides a clear framework for understanding how FMs contribute to different stages of anomaly detection.
		\item \textbf{A systematic review.} We conduct an extensive review of state-of-the-art methods that leverage FMs for anomaly detection, organizing them according to our proposed taxonomy. This review highlights the latest trends, methodologies, and applications across various domains.

		\item \textbf{Future directions.} We identify key challenges in FM-based anomaly detection, including efficiency, bias, explainability, and multimodality. Furthermore, we outline promising future research directions to address these challenges and advance the field.

	\end{itemize}

	\section{Preliminaries}
	We introduce some basic concepts of FMs and present our proposed taxonomy of anomaly detection using FMs.
    
	\subsection{Foundation Models}

	\textbf{Definition.} FMs are a class of large, pre-trained models that serve as the common basis for a wide range of downstream tasks across various domains. First introduced by~\citep{bommasani2021opportunities}, FMs build on concepts from deep neural networks and self-supervised learning. A prominent subset of FMs, LLMs, are trained on extensive text datasets and are able to perform tasks like language generation, comprehension, and translation. Unlike LLMs, which are limited to written language, FMs have a broader scope, capable of processing diverse data types such as text, images, and audio. They undergo large-scale pre-training on vast datasets and can be fine-tuned with task-specific data, making them highly versatile for numerous applications.
	
	\textbf{Composition.}
	FMs are usually built on the basis of large data, self-supervised pretraining, and transformer-based architecture~\citep{zhou2024comprehensive}. Pretraining focuses on training a general model to learn generic representations using large amounts of diverse, unlabelled data. This general model, inspired by transfer learning~\citep{niu2020decade}, could then be used to perform different downstream tasks and enhance model performance in other fields through fine-tuning. In the pretraining stage, self-supervised learning~\citep{liu2022graph} is applied across a wide range of domains and areas where unlabeled data is naturally available. As for the model structure, the transformer architecture is the most popular for FMs in different areas, like natural language processing (NLP) and computer vision (CV). These pretrained models may be adapted to specific tasks using a number of methods, such as transfer learning to fine-tune some or all of the parameters of FMs with a much smaller task-specific dataset~\citep{zhang2022transfer}, few-shot learning with only a few samples, or zero-shot learning without any task-specific examples.
 \begin{figure*}
\centerline{\includegraphics[width=0.9\linewidth]{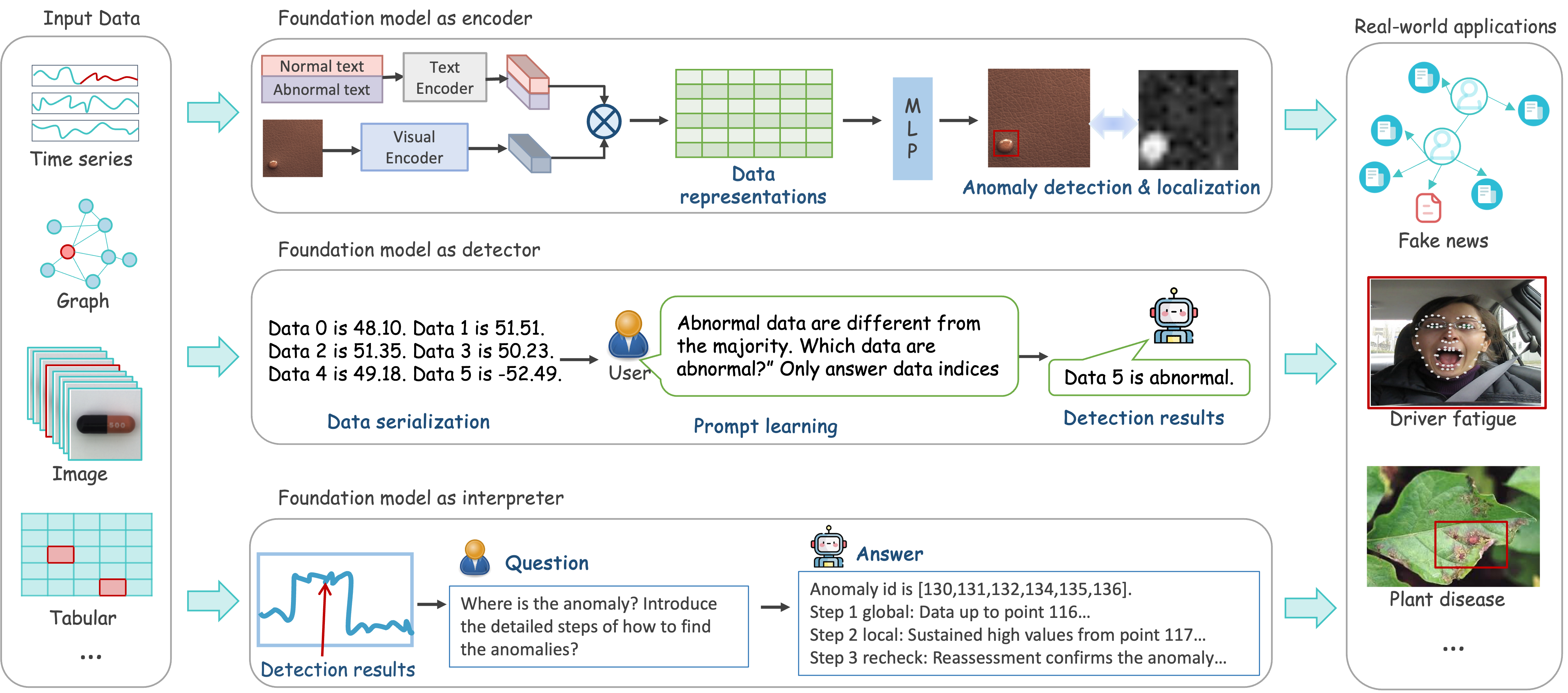}}	
\caption{An illustration of applying foundation models to the three key stages of anomaly detection, playing roles as encoder, detector, and interpreter. Three cases are illustrated with image, tabular, and time series data, respectively.}
\label{fig:framework001}
\end{figure*}   

	\subsection{Proposed Taxonomy}
	As shown in Figure~\ref{fig:framework001}, the taxonomy of this survey classifies current anomaly detection FMs into three main categories: 1) FM as Encoder, where FMs are incorporated into the process of computing embeddings for high-level representations; 2) FM as Detector, where FMs are directly used as anomaly detectors to identify and localize specific anomalies; 3) FM as Interpreter, where FMs are assisted in providing explanations on the causes of anomalies. Examples of downstream tasks include fake news detection in social networks, driver fatigue detection in autonomous robotic systems, and plant disease detection in agricultural systems, to name a few.

	It should be noted that in some works, due to the diversity of FMs, more than one kind of FMs are used, and sometimes they serve diverse roles in different stages of anomaly detection. Therefore, it is inappropriate to categorize them exclusively into one of these three main classes. For example, LogiCode~\citep{zhang2024logicode} transforms structured request prompts and logical rules into executable Python codes by harnessing the power of the LLM in this regard, which regards the foundation model as a code generator. AnomalyRuler~\citep{yang2025follow} employs two different FMs, a Vision-Language Model (VLM) to generate descriptions for input video frames and a LLM to derive rules for anomaly detection by contrasting the rules for normality. Audit-LLM~\citep{song2024audit} regards a FM as an assistant by serving as task decomposer, tool builder, and executor in every stage during the whole process of insider threat detection.
    
	Table~\ref{sum} summarizes the models that leverage FMs to assist anomaly detection tasks according to the proposed taxonomy. In the next three sections, we will systematically examine the state of the art in this field, organizing our review by category. 
	
	\section{FM as Encoder}
\begin{figure}
		\centering  
		\vspace{-0.35cm} 
		\subfigtopskip=2pt 
		\subfigbottomskip=2pt 
		\subfigcapskip=-5pt 
		\subfigure[FM-based Embedding]{
			\label{fig:fmena}
			\includegraphics[width=0.39\linewidth]{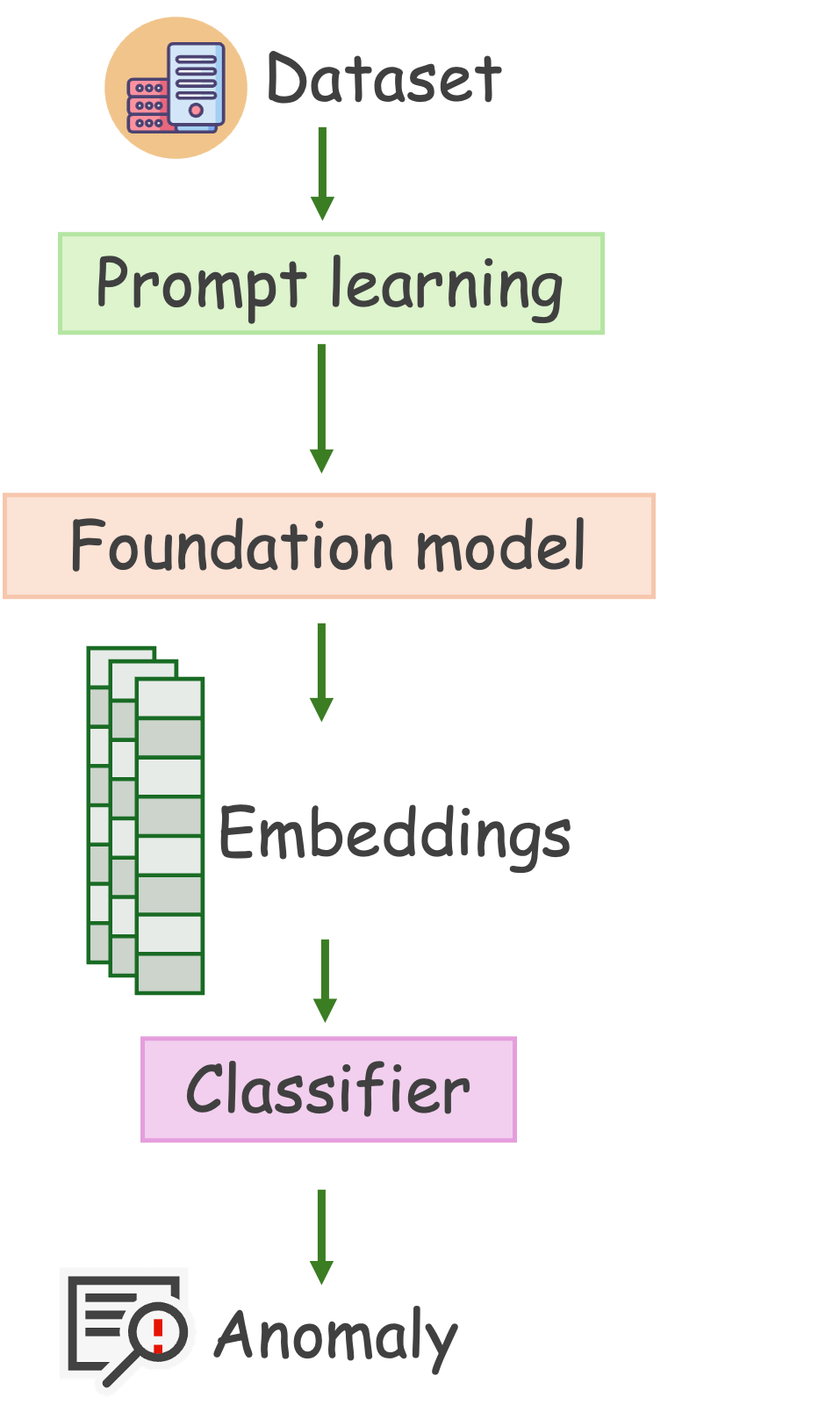}}
		\quad 
		\subfigure[Hybrid Embedding]{
			\label{fig:fmenb}
			\includegraphics[width=0.36\linewidth]{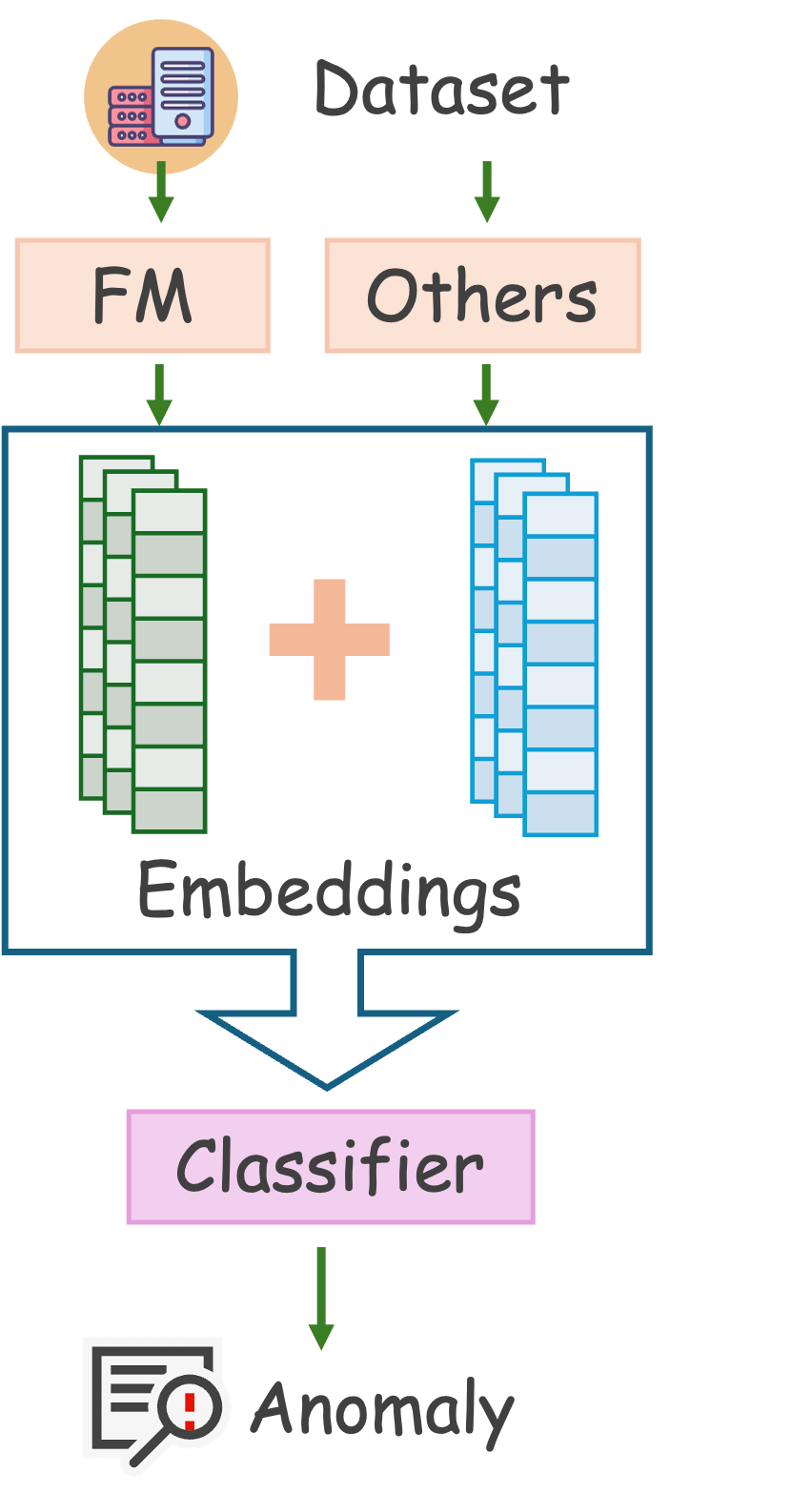}}
		
		\caption{FM as encoder.}
		\label{fig:fmen}
	\end{figure}

	In an anomaly detection model, the encoder module plays a critical role in transforming input data (e.g., text, time series, images, etc.) into a latent feature embeddings. This representation captures essential characteristics of the data samples that can later be analyzed to detect deviations or anomalies. FM as encoder approaches focus on enhancing the quality of data embeddings with the help of powerful FMs.
	The derived embeddings are directly inputted into downstream
	classifiers for anomaly detection. Referring to Figure~\ref{fig:fmen}, we naturally categorize these approaches into two branches: FM-based and hybrid embedding, depending on whether or not the latent embeddings are only generated by the foundation model.
	
	There is a special case where FM is not the encoder of data samples. We classify this kind of models in this section because the FM indirectly participates in the encoding process. Specifically, AnomalyLLM~\citep{liu2024large} is a knowledge distillation-based time series anomaly detection model by employing the foundation model as its teacher network. The anomalies are detected based on the discrepancy between the features of the teacher and student networks, where a large representation gap reflects anomalous samples. 
	
	\subsection{FM-based Embedding}
	With reference to Figure~\ref{fig:fmena}, FMs are directly utilized as the unique encoder to output embeddings into the classifier for anomaly detection:
		\begin{align}
			\textnormal{Embedding:}&~~ \textbf{z}_i=\phi_{FM}(x_i),\\
			\textnormal{Anomaly Detection:} &~~~\hat{Y}~=\tau(\textbf{z}_i).
		\end{align}
	In these equations, the text information $x_i$ is encoded by a foundation model $\phi_{FM}(\cdot)$ into $\textbf{z}_i$. The embeddings generated by this kind of approach are directly fed into the classifier $\tau(\cdot)$ for classification label $\hat{Y}$, typically without the need for any other encoders. Generally, most of these works use a Large Vision-Language Model (LVLM) to generate both textual and visual embeddings for image/video anomaly detection.
	
	In the case of image anomaly detection, there are generally two encoders, one for text embedding to describe the content of images, the other for visual embedding to encode images~\citep{radford2021learning}. One representative example is WinCLIP~\citep{jeong2023winclip}, which proposed language-guided zero-shot anomaly detection by introducing a compositional prompt ensemble. To leverage the normal images available in the few normal shots setting, its few-normal-shot extension WinCLIP+ is introduced to aggregate complementary information from WinCLIP and visual signals from normal samples. However, the performance of WinCLIP+ is heavily dependent on extensive engineering on hundreds of manually defined prompts. To solve this problem, AnomalyCLIP~\citep{zhou2024anomalyclip} uses an object-agnostic prompt template to model the semantics of general abnormality and normality, thus improving the performance of generalized zero-shot anomaly detection.  A similar model, InCTRL~\citep{zhu2024toward}, also explores the problem of generalist anomaly detection, meaning that a single model can detect anomalies in different data sets from various application scenarios without the need of additional training on the target dataset. The detection strategy is to identify the discrepancies between query images and a set of few-shot normal images from the auxiliary data, where anomalies are expected to have larger discrepancies than normal samples. Moreover, ALFA~\citep{zhu2024llms} and MVFA \citep{huang2024adapting} propose to solve generalist anomaly detection in zero/few-shot scenarios as well. Specifically, a run-time prompt adaptation strategy is utilized in ALFA to generate informative anomaly prompts for every image. Then, a fine-grained aligner is developed to learn local semantic space projection, which can be generalized to support pixel-level anomaly localization. As for MVFA, the authors design a multi-level visual feature adaptation architecture for medical image anomaly detection, on the basis of CLIP model. In \citep{sinha2024real}, the authors use relatively small FMs (e.g., 120M parameters) to detect deviations from prior experiences in real time. At the same time, the LLM-based monitor can reason about the safety consequences of anomalous scenarios, thereby determining whether intervention is necessary.
	
	\subsection{Hybrid Embedding}
	Approaches using hybrid embeddings focus on utilizing the capability of FMs to capture additional information as shown in Figure~\ref{fig:fmenb}. Taking graph anomaly detection as an example, the embeddings are generated by combining the output of both foundation model $\phi_{FM}(cdot)$ and graph neural networks (GNNs) $\phi_{GNNs}(cdot)$:
		\begin{align}
			\textnormal{Embedding:}&~\textbf{z}_i=F(\phi_{FM}(x_i), \phi_{GNN}(x_i)),\\
			\textnormal{Anomaly Detection:} &~\hat{Y}=\tau(\textbf{z}_i).
		\end{align}
	There are multiple fusion strategies $F(\cdot)$ of embeddings, such as concatenation, weighted averaging, or using attention mechanisms to highlight relevant features. This fused representation captures both semantic and relational information, enriching the dataset without the need for large quantities of labeled data.
	
	For example, AnomalyLLM~\citep{liu2024anomalyllm} was proposed to detect anomaly edges by aligning LLM with dynamic graphs. Instead of directly using LLM as an encoder, the backbone LLM takes the embedding from a GNN encoder as input to generate the final representation vector for anomaly detection. Despite that FMs have the powerful capacity to capture semantic information,
	\citep{bakumenko2024advancing} uses LLMs to encode non-semantic categorical data (i.e., journal entries) from real-world financial records. Compared with traditional encoders, LLMs could better solve the issues of feature heterogeneity and sparsity in financial audits. In \citep{bakumenko2024advancing}, a hybrid model that combines sentence-transformer embeddings with machine learning (ML) classifiers is designed to enhance anomaly detection performance. To compare the performance of different models, three LLMs and five ML classifiers are evaluated from the perspectives of quality, efficiency, and speed, thereby facilitating a comprehensive evaluation.

	\textbf{Discussion.}
	Utilizing FMs as encoder has demonstrated superior performance on anomaly detection, being capable of effectively capturing rich semantic, general-purpose patterns. At the same time, many multimodal FMs (e.g., CLIP or GPT-4) enable joint processing of data from multiple modalities (e.g., text, image, video) into a unified space, enhancing anomaly detection in complex and multimodal datasets. Moreover, FMs are more resilient to imperfect data thanks to its pretraining stage. However, despite some papers claiming strong scalability~\citep{zhang2025scalalog}, the resource-intensive characteristic of FMs limits their use in real-time or edge-based anomaly detection applications. Besides, the representations generated by FM directly impact the detection results of models, which may in turn pose transparency and trustworthiness issues in critical applications. 
		
	\begin{figure}
		\centering  
		\vspace{-0.35cm} 
		\subfigtopskip=2pt 
		\subfigbottomskip=2pt 
		\subfigcapskip=-5pt 
		\subfigure[Serialization-based]{
			\label{fmdea}
			\includegraphics[width=0.36\linewidth]{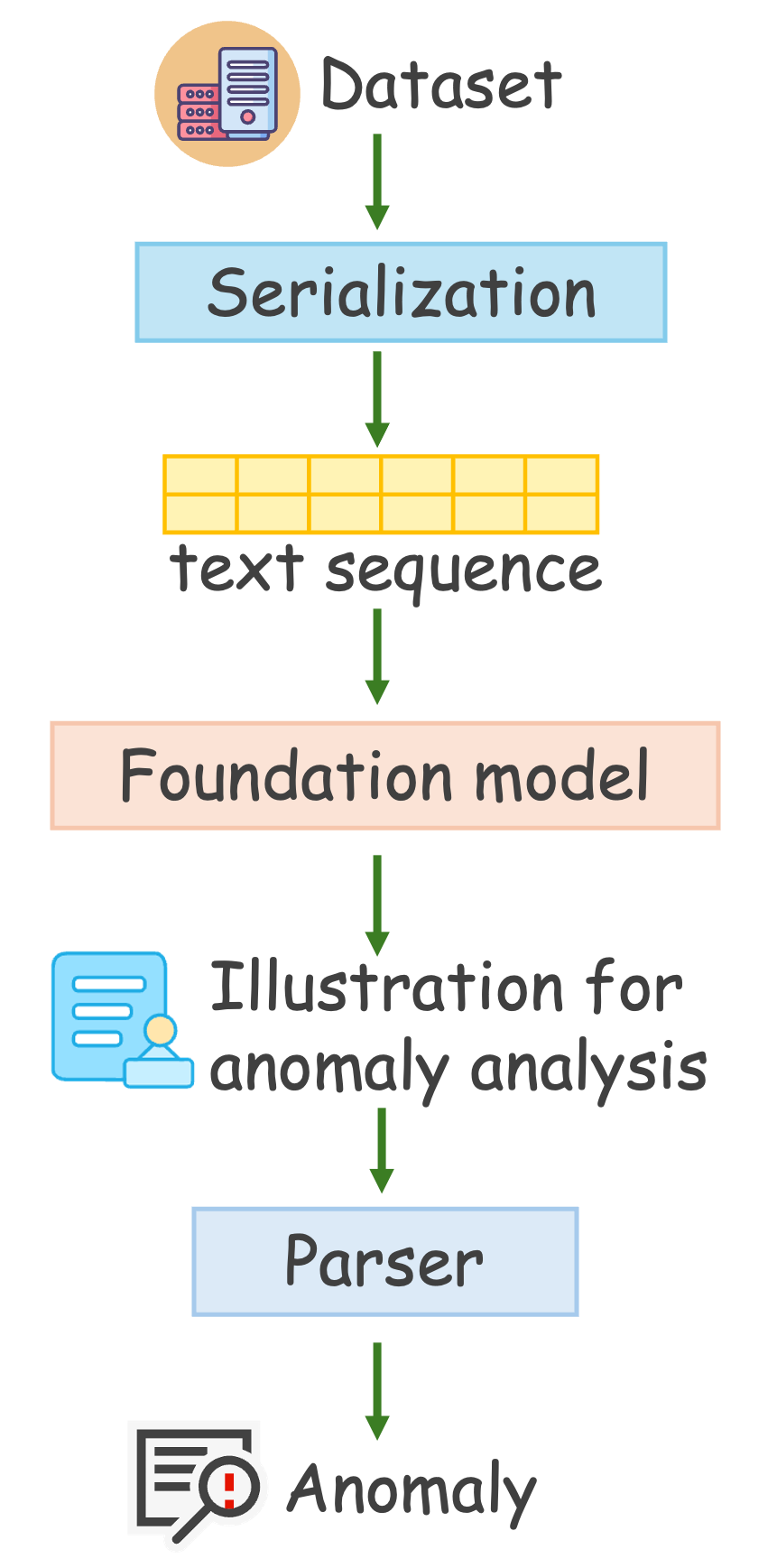}}
		\quad 
		\subfigure[Encoding-based]{
			\label{fmdeb}
			\includegraphics[width=0.37\linewidth]{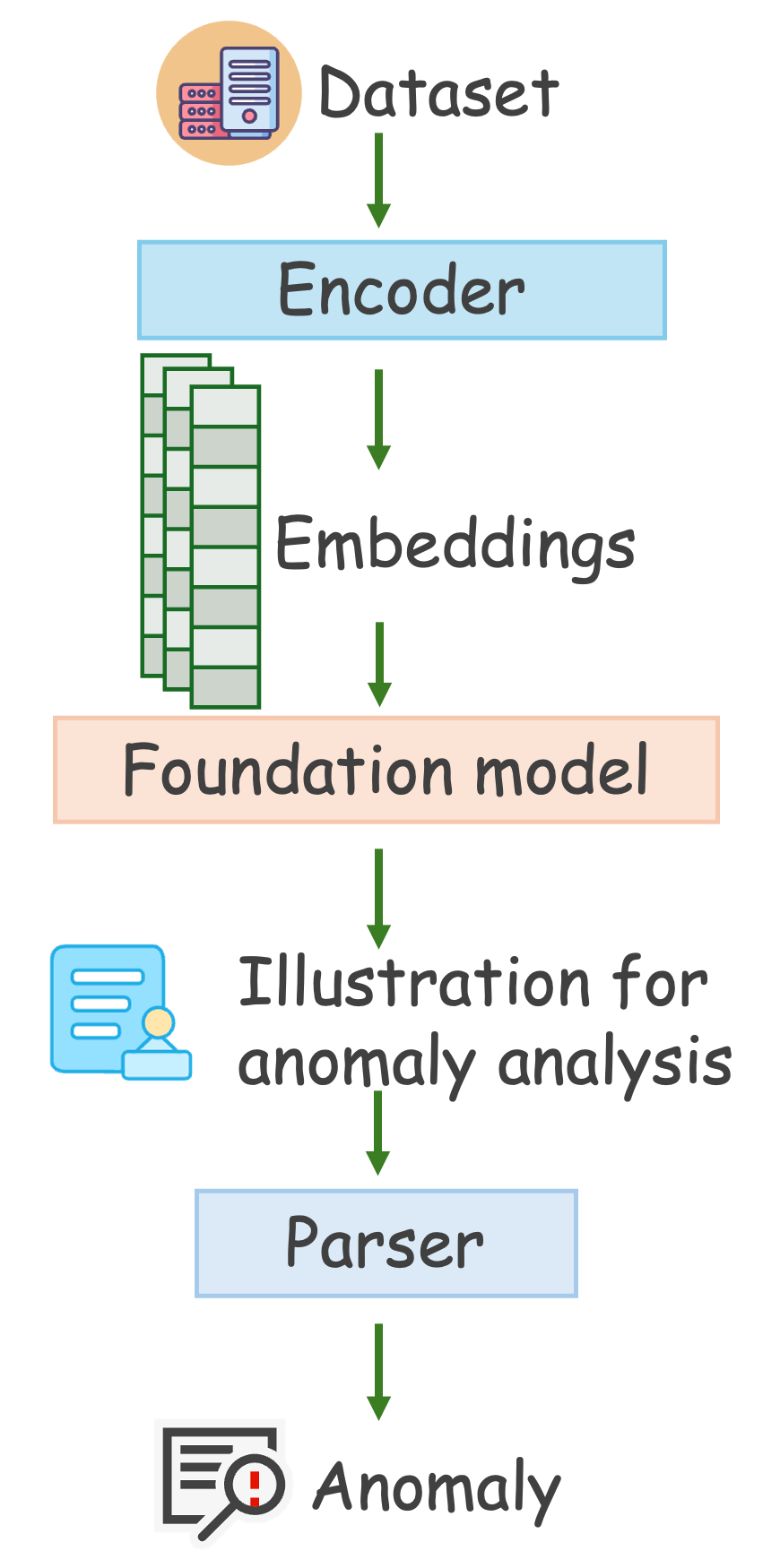}}
		
		\caption{FM as detector}
		\label{fmde}
	\end{figure}
	\section{FM as Detector}
	The core idea behind this category is to utilize FMs as anomaly detectors to detect anomalies for a wide range of tasks, including classifications and localization. However, applying FMs directly as detectors presents unique challenges, primarily because FMs are often prompt-based, while the source data in most cases present different modalities, such as time-series, image, and video. In this section, we classify the models broadly into serialization-based and encoding-based detection, depending on how data are preprocessed before inputted to FMs.
	\subsection{Serialization-based Detection}
	Most of the existing research attempts to employ the question-answering capability of the FM to directly detect anomalies, which omits the representation learning process of data embeddings. As shown in Figure \ref{fmdea}, serialization-based detection typically involves two steps: (1) transforming source data into a sequence of text with a serialization function $SRL(\cdot)$, and (2) extracting the detected anomalies from the FM output with a parsing function $Parse(\cdot)$, as illustrated below: 
		\begin{align}
			\textnormal{Data Serialization: }& \mathcal{X}_{txt} = SRL(\mathcal{X}),\\
			\textnormal{Prediction: }& \hat{Y} = Parse(\phi_{FM}(\mathcal{X}_{txt},p)),
		\end{align}
	where $p$ denotes the instruction prompt for the specific task.
	
	The parsing strategies of models are generally standardized. For example, given that the output of FMs often involves their reasoning and logic processes, several works~\citep{dong2024can,zanella2024harnessing,li2024anomaly} utilize specific prompts to extract the predicted label from the output. Another way is to regard anomaly detection as a Q\&A problem where FMs are instructed to answer questions in a specific format. For instance, some works~\citep{gu2024anomalygpt,alnegheimish2024large,elhafsi2023semantic,chaudhuri2024spiced} limit LLM’s output format via guiding instructions in prompts, such as “This is a photo of leather. Is there any anomaly in the image?”. In addition, some methods~\citep{hadadi2024anomaly} fine-tune FMs to directly output the label of anomalies, allowing them
	to provide accurate predictions without additional parsing steps.
	
	Considering that FMs sometimes show failure cases, such as pairing incorrect indices and values, generating indices beyond the batch length, and listing every data item as abnormal, Li et al.~\citep{li2024anomaly} simulate a synthetic dataset with ground-truth labels for LLMs to be fine-tuned in a supervised manner. Note that the data are serialized into text before being inputted into the FMs. When detecting anomalies from unstable logs in software systems, one critical challenge is the lack of information about new logs because there is insufficient log data in new software versions. To mitigate the data insufficiency issue,~\citep{hadadi2024anomaly} pre-trains LLMs on vast amount of data for robust understanding of diverse patterns and contextual information. Specifically, the authors compare fine-tuned and prompt-engineered FMs to demonstrate which strategy performs better. 
	
	SIGLLM~\citep{alnegheimish2024large} is mainly made up of two parts, a PROMPTER to identify parts of a sequence that LLM thinks are anomalous and a DETECTOR to find anomalies using the residual between the original signal and the forecast. Before inputting into the LLM, a time series-to-text conversion module is devised to convert time series data into LLM-ready input. Similarly, to explore whether FMs could be used as a time series anomaly detector, Dong et al.~\citep{dong2024can} design a prompt learning strategy and synthesize a data set to automatically generate time series anomalies with explanations. With instruction fine-tuning on this dataset, the authors demonstrate that foundation models can yield improved performance in time-series anomaly detection tasks.
	
	Elhafsi et al.~\citep{elhafsi2023semantic} introduce a monitoring framework to detect semantic anomalies limited by vision-based policies; the monitor is composed of an LLM to infer what kind of observed objects in a scene may lead to confusion that could result in policy errors. To improve the reasoning ability of LLM, prompt engineering like few-shot prompting and chain-of-thought reasoning are employed in the instantiations of this framework. The same prompt learning strategy is used in SPICED~\citep{chaudhuri2024spiced}, which is an LLM-based framework for the detection and localization of syntactical bugs and analog Trojans in circuit net lists.
	
	\begin{table*}\tiny
		\caption{A summary of models that leverage FMs to assist anomaly detection or prediction tasks in literature. \textbf{FMs} shows the specific open-sourced FMs used in the paper. Acronyms in FM type: LLM (Large Language Model); LVLM (Large Vision-Language Models); MLLM (Multimodal Large Language Model). \textbf{Fine-tuning} denotes whether the parameters of FMs are fine-tuned during the detection process, and $\bigstar$ indicates that models employ parameter-efficient fine-tuning (PEFT). \textbf{Prompting} indicates the use of text-formatted prompts in LLM. \textbf{Data} indicates the anomalous data type detected in the paper. } \label{sum}
		\begin{tabular}{p{0.01cm}<{\centering} p{2cm}<{} p{1.5cm}<{} p{1cm}<{\centering}  p{1.2cm}<{\centering}  p{1cm}<{\centering} p{1cm}<{} p{2cm}<{} p{0.5cm}<{\centering}}
			
			\hline
			&	\textbf{Model}&\textbf{FMs}&\textbf{FM type}&\textbf{Fine-tuning}&\textbf{Prompting}&\textbf{Data}&\textbf{Domain}&\textbf{Code}\\
			\hline
			\multirow{8}{*}{\rotatebox{90}{FM as Encoder}}			&	
			\citep{bakumenko2024advancing}	& Transformer &LLM &\XSolidBrush&\XSolidBrush&Tabular&Finance&-	\\
			
			&	ANOMALYLLM\citep{liu2024large}&GPT-2 &LLM	&\Checkmark&\XSolidBrush&Time series&-&-\\
			&ALFA \citep{zhu2024llms}	& GPT &LVLM &\XSolidBrush&\Checkmark&Image&Industry&	-\\
			&InCTRL \citep{zhu2024toward}	& OpenCLIP &LVLM &\XSolidBrush&\Checkmark&Image&-&	\href{https://github.com/mala-lab/InCTRL}{Link}\\
			&AnomalyCLIP \citep{zhou2024anomalyclip}	& CLIP &LVLM &\Checkmark&\Checkmark&Image&-&\href{https://github.com/zqhang/AnomalyCLIP}{Link}	\\
			
			&WinCLIP \citep{jeong2023winclip}	& OpenCLIP &LVLM &\XSolidBrush&\Checkmark&Image&-&- \\
			&MVFA \citep{huang2024adapting}	& CLIP &LVLM &\Checkmark&\Checkmark&Image&Medical&\href{https://github.com/MediaBrain-SJTU/MVFA-AD}{Link} \\
			&	\citep{sinha2024real}&Many &LLM	&\XSolidBrush&\Checkmark&Text&Robotics&\href{https://sites.google.com/view/aesop-llm}{Link}\\
			& AnomalyLLM \citep{liu2024anomalyllm}&Transformer &LLM	&\XSolidBrush&\Checkmark&Graph&-&\href{https://github.com/AnomalyLLM/AnomalyLLM}{Link}\\
			&\citep{kim2023unsupervised}	& ChatGPT &LVLM &\Checkmark&\Checkmark&Video&Real-world surveillance&	-\\
			\hline
			\multirow{9}{*}{\rotatebox{90}{FM as Detector}}	&LAVAD~ \citep{zanella2024harnessing}	& Llama-2&LVLM\&LLM &\XSolidBrush&\Checkmark&Video&Real-world surveillance&	\href{https://lucazanella.github.io/lavad/}{Link}\\
			&AnomalyGPT \citep{gu2024anomalygpt}	& Vicuna &LLM &\Checkmark&\Checkmark&Image&Industry&-	\\
			
			&\citep{li2024anomaly}	& GPT-4 &LLM &~~~~~${\text{\Checkmark}}^{\bigstar}$
			&\Checkmark&Tabular&- &	-\\
			&\citep{elhafsi2023semantic}	& - &LLM &\XSolidBrush&\Checkmark&Video& Autonomous driving&-	\\
			
			&\citep{dong2024can}	& GPT-4 \& LLaMA3 &LLM &~~~~~${\text{\Checkmark}}^{\bigstar}$
			&\Checkmark&Time series&-&-	\\
			&	LLMAD \citep{liu2024large2}&GPT-4 &LLM	&\XSolidBrush&\Checkmark&Time series&-&-\\
			&SIGLLM \citep{alnegheimish2024large}	& GPT \& MISTRAL &LLM &\Checkmark&\Checkmark&Time series&-&\href{https://github.com/sintel-dev/sigllm}{Link}	\\
			&\citep{hadadi2024anomaly}	& GPT-3 &LLM &\Checkmark&\Checkmark&Log&Software system&	-\\
			&SPICED~\citep{chaudhuri2024spiced}	& GPT-3.5 &LLM &\XSolidBrush&\Checkmark&Signal& Electronics&	-\\
			
			\hline
			
			\multirow{5}{*}{\rotatebox{90}{Interpreter}}		
			&	
			Holmes-VAD \citep{zhang2024holmes}& Llama3-Instruct-70B&MLLM	&~~~~~${\text{\Checkmark}}^{\bigstar}$
			&\Checkmark&Video&Real-world surveillance&\href{https://holmesvad.github.io/}{Link}\\		
			&\citep{park2024enhancing}	& - &LLM &\XSolidBrush&\Checkmark&Tabular&Finance&	-\\
			&	DABL~\citep{guan2025dabl}&Llama-2 &LLMs	&~~~~~${\text{\Checkmark}}^{\bigstar}$
			&\Checkmark&Text&Business&\href{https://github.com/guanwei49/DABL}{Link}\\
			&Myriad~\citep{li2023myriad}	& GPT-3.5 &LVLM &\XSolidBrush&\Checkmark&Image& Industry&\href{https://github.com/tzjtatata/Myriad}{Link}	\\
			&	VAD-LLaMA~\citep{lv2024video}&LLaMA &LVLM	&\XSolidBrush&\Checkmark&Video&Real-world surveillance&-\\
			&	LogConfigLocalizer~\citep{shan2024face}	& GPT-4 &LLM &\XSolidBrush&\Checkmark&Log&Software system&	\href{https://github.com/shanshw/LogConfigLocalizer/}{Link}\\

			\hline
			\multirow{3}{*}{\rotatebox{90}{Others}}			
			&LogiCode \citep{zhang2024logicode}	& GPT-4&LLM &\XSolidBrush&\Checkmark&Image&Industry	&-\\
			&AnomalyRuler~\citep{yang2025follow}	& GPT-4 \& Mistral-7B &LVLM\&LLM &\XSolidBrush&\Checkmark&Video&Real-world surveillance&\href{https://github.com/Yuchen413/AnomalyRuler}{Link}\\
			&Audit-LLM~\citep{song2024audit}&GPT-3.5 &LLM	&\XSolidBrush&\Checkmark&Log&Audits&-\\
			
			\hline
		\end{tabular}
	\end{table*}
	\subsection{Encoding-based Detection}
	Deep learning (DL) has demonstrated impressive capabilities in understanding different data structures through neural networks, which excel at recognizing patterns in complex and high-dimensional data. As illustrated in Figure
	\ref{fmdeb}, encoding-based detection leverages the advantages of deep learning to incorporate complex characteristics present in source data, allowing FMs to be characteristic-aware: 
		\begin{align}
			\textnormal{Representation Learning:}& ~~~\textbf{z}_i = f_{DL}(x_i),\\
			\textnormal{Prediction:}& ~~~\hat{Y} = Parse(\phi_{FM}(\textbf{z}_i,p)),
		\end{align}
	where $f_{DL}$ is the encoder for data representation learning based on deep learning. DL-based prediction also utilizes a parser to retrieve FM output.
	
	To combat the challenge that traditional industrial anomaly detection models can only provide anomaly scores of data samples and the threshold must be manually determined, AnomalyGPT~\citep{gu2024anomalygpt} employs an image decoder to provide fine-grained semantics and introduces a prompt learner to fine-tune the LVLM using prompt embeddings. Therefore, it can directly assess the locations of anomalies and provide image information in a human-understandable manner.

	LAVAD \citep{zanella2024harnessing} uses LLMs to detect video anomalies exclusively from a scene description without further training. After generating a textual description for each video frame with a pre-trained VLM, an LLM is applied to capture the dynamics of the scene and summarize captions within a temporal window, which will be further used to provide an anomaly score for each frame.

	\textbf{Discussion.} 
	Utilizing FMs directly as detectors shows superiority in
	processing textual attributes of different types of data, especially achieving remarkable zero/few-shot performance compared with traditional DL models. The ultimate goal is to develop and refine methods for encoding these structured information into a format that
	FMs can comprehend and manipulate effectively and efficiently. Despite the fact that LLMs have shown their potential in understanding long-context and mathematical reasoning, they cannot be directly applied as an anomaly detector without prompt learning and fine-tuning strategy. Moreover, FMs tend to miss or normalize anomalies, especially some subtle or domain-specific ones, because they are trained on large corpora of normal data without specific training stages~\citep{russell2024aad}.
    As for encoding-based detection, training an additional DL module and inserting it into FMs for joint training is challenging due to the problem of vanishing gradients in the early layers of deep transformers~\citep{li2024survey}.
	
	\begin{figure}
		\centering  
		\vspace{-0.35cm} 

		\subfigure[Direct detection]{
			\label{fmina}
			\includegraphics[width=0.3\linewidth]{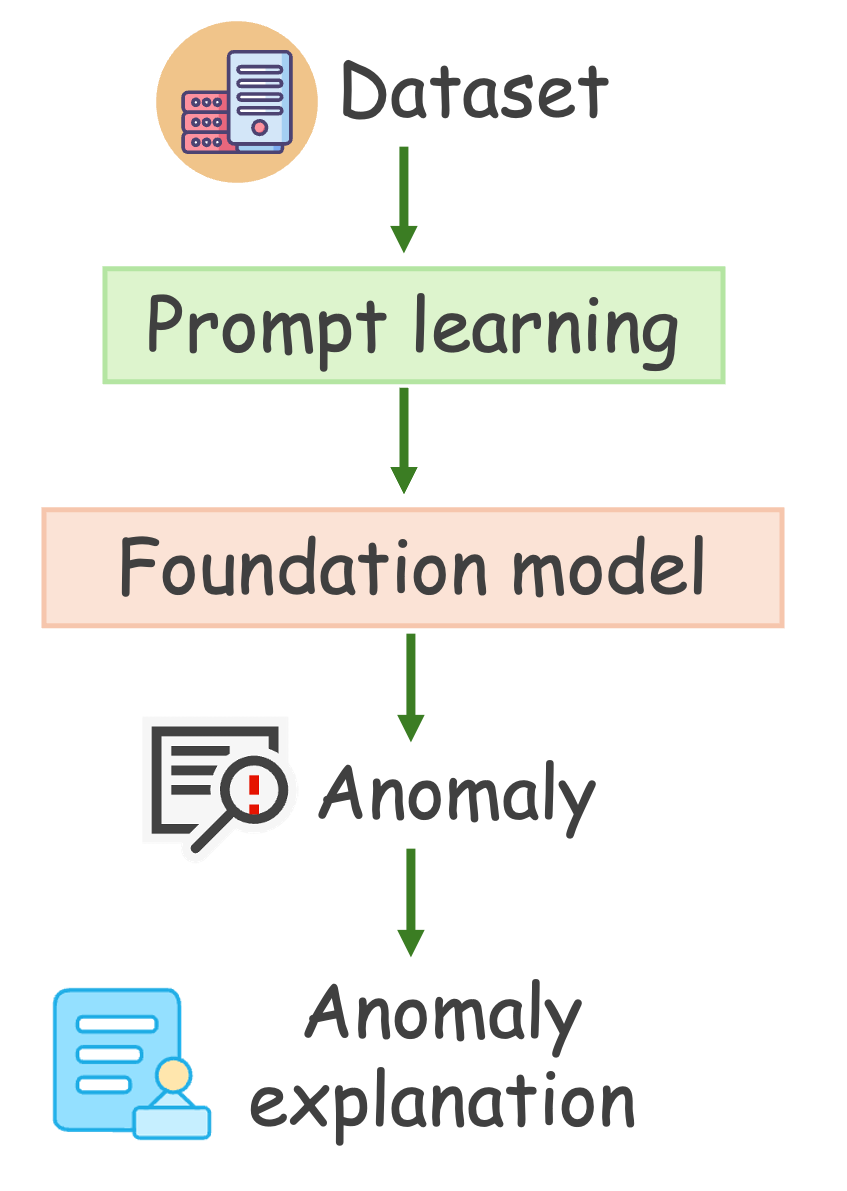}}
		\subfigure[Indirect detection]{
			\label{fminb}
			\includegraphics[width=0.3\linewidth]{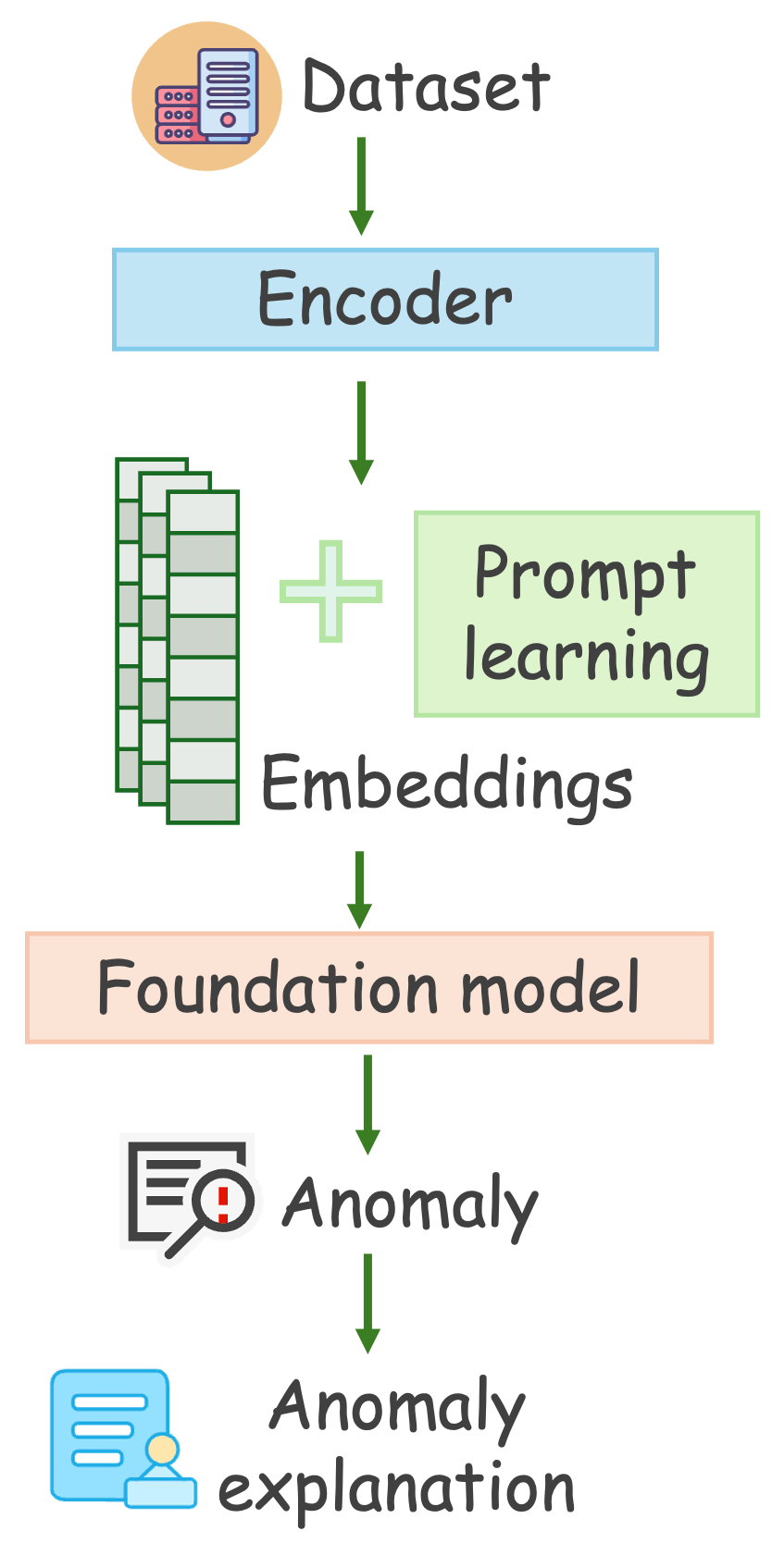}}
		\subfigure[Verification]{
			\label{fminc}
			\includegraphics[width=0.3\linewidth]{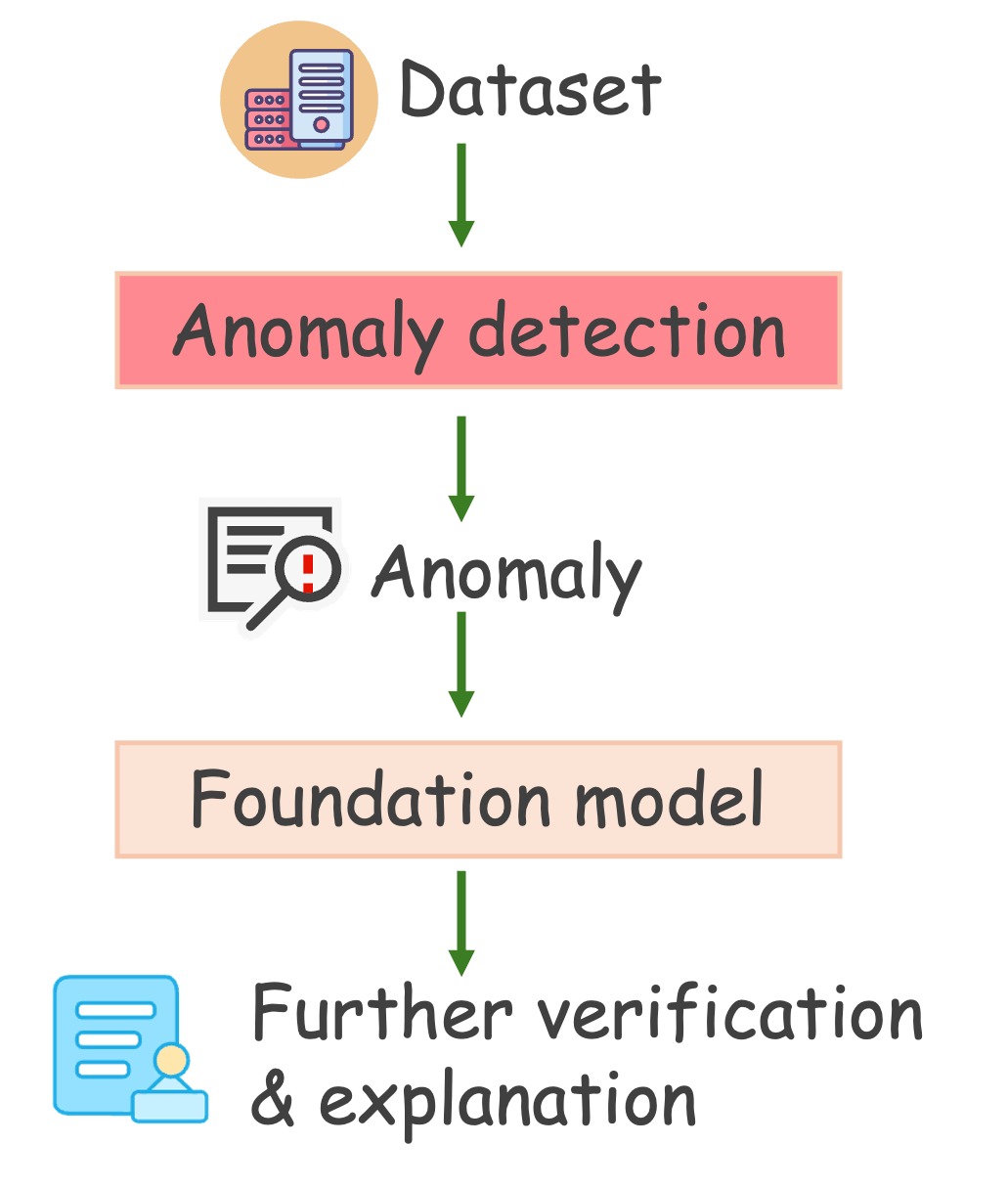}}
		\caption{FM as interpreter}
		\label{fmde}
	\end{figure}
	
	\section{FM as Interpreter}
	In the last stage of anomaly detection, the FMs could be used as an interpreter to provide explanations on the detection results. We classify the models into detection-based and verification-based models based on whether FM serves as an anomaly detector as well.
	\subsection{Detection-based Explanation}
	This classification is based on the primary focus of interpreting anomalous samples, including the reasoning behind anomalies and the additional context they provide. As illustrated in Figures~\ref{fmina} and \ref{fminb}, these approaches can be grouped into direct and indirect detection, depending on the input structure used with FMs. 
	Correspondingly, the explanation $Exp$ of FMs $f_{FM}(\cdot)$ is paired with a corresponding anomaly-aware question $P_d$ to construct an instruction item set $\mathcal{I}$:
		\begin{align}
			&\textnormal{Explanation:} ~~Exp = f_{FM}(P_t,y_i,c_i)\\
			&\textnormal{Instruction Dataset Construction:}\notag\\
			&\mathcal{I}_i = \{``user:":P_d, ``FM": Exp\},
		\end{align}
	where task prompt $P_t$ combined with the abnormal label $y_i$ and detailed caption $c_i$ are inputs to the foundation model $f_{FM}(\cdot)$ to make judgements and provide explanations.
	
	\textbf{Direct detection.}
	Holmes-VAD \citep{zhang2024holmes} is composed of three key components, namely, a Video Encoder for encoding input video, a Temporal Sampler for abnormal frame prediction, and a Multi-modal LLM for generating text explanations. To create a generic model capable of detecting semantic anomalies in business processes, DABL~\citep{guan2025dabl} fine-tunes the Llama by incorporating traces into question and answer content. Specifically, every trace item (i.e., business process) and its label are inputs to the foundation model so that the model could be fine-tuned in a supervised way. Then, the fine-tuned model outputs whether the given trace is normal or anomalous, and provides causes of anomalous traces. LLMAD~\citep{liu2024large2} is an LLM-based framework for time series anomaly detection. It improves detection performance and interpretation quality by injecting data background and domain knowledge into model via time-series In-Context Learning and Chain-of-Thought approach guiding its decision-making process. 
	
	\textbf{Indirect detection.}
	In Myriad~\citep{li2023myriad}, a vision expert tokenizer embeds the anomaly map into vision expert tokens to make LLM perceive prior knowledge, and a vision expert instructor generates domain vision-language tokens to compensate for the errors of vision experts. VAD-LLaMA~\citep{lv2024video} is composed of a video encoder and a foundation model, which are connected by a projection layer called adapter. By treating the foundation model as a question-answer system, the foundation model provides detailed information (e.g., the concrete video content and the accurate time frames when anomalies appeared) of anomalies by answering the human queries.
	
	\subsection{Verification-based Explanation}
	Verification-based FMs aim to further verify the detection results by providing some textual explanations about the detected anomalies (Figure~\ref{fminc}). To ensure the effectiveness of FMs, specific prompts are designed to consistently generate identical output for the same queries.
	
	To further validate the authenticity of detected anomalies, ~\citep{park2024enhancing} proposes an LLM-based multi-agent framework to analyze the detailed anomaly information, which serves as a critical interface between the AI-driven analysis process and human decision-making.
	With the aim of assisting end-users in coping with configuration errors in software systems through log analysis, ~\citep{shan2024face} proposes a two-stage strategy to localize the root-cause configuration logs. The first stage is to identify anomalous logs through obtaining key log messages, and the second stage is to use an LLM for further verification with the help of their strong power in natural language understanding and processing. Specifically, strategies are designed to increase the reliability of LLMs’ judgments and provide additional information about the configuration errors.
	
	\textbf{Discussions.}
It is acknowledged that providing human-understanable explanations for anomaly detection results is useful for high-stake domains like healthcare and finance, where domain experts without AI knowledge can engage with the system.	Apart from the explanation results, incorporating external knowledge that are not encoded in the raw data has the potential to reduce false positives through context~\citep{zhang2024logicode}.

Although FMs excel in providing explanations on the anomaly detection results, its interpretability is still limited due to the difficulty of well-crafted prompts. Actually, creating these prompts needs domain specific knowledge from experts~\citep{cao2024domain}. Therefore, how to improve the data quality with acceptable labeling costs is still a core challenge to be solved in the future.
	\section{Open Challenges and Way Forward}
	Based on the above review and analysis, we believe that there are still many open challenges and potentials for further advancements in this field. In this section, we list some future research
	directions for further exploration.
	
	\subsection{Efficiency} 
	Notwithstanding the stellar progress and accomplishments of FMs, the improvements of performance in these models come at the expense of the model's efficiency. However, anomaly detection models need to be particularly efficient because they are often applied in real-time, high-stakes environments where timely detection and response to anomalies are critical. Common application scenarios include fraud detection and healthcare, where even small delays of identification may lead to significant financial loss or even safety risks. While some lightweight adaptation techniques have been proposed to improve the parameter efficiency, Lester et al.~\citep{lester2021power} give an example that, as the model size increases, the performance gap between full fine-tuning and lightweight adaptation is diminishing rapidly.
	Therefore, exploring mechanisms to optimize the trade-off between efficiency and expressivity of FMs still remains a notable challenge. 
	\subsection{Bias} 
	In recent years, FMs have led to an extraordinary level of homogenization, which refers to the consolidation of methodologies for building machine learning systems across a wide range of applications~\citep{bommasani2021opportunities}: current NLP models are often adapted from one of popular FMs like BERT. Despite that any improvements of FMs can help produce immediate benefits, such kind of extension might also have the potential to inherit or even amplify the problematic biases of these models. A biased anomaly detection model may potentially lead to incorrect or unfair outcomes. For example, a biased model may disproportionately flag specific data points or groups (e.g., gender and racial bias) as anomalous even when they are normal, limiting their capability of detecting genuine anomalies. Such kind of false positives or false negatives can lead to unfair treatment of certain individuals or groups, missed critical alerts, or a lack of trust in the system, ultimately compromising its effectiveness and fairness.
	Therefore, it is critical to eliminate the intrinsic bias present within FMs, thereby further avoiding extrinsic harms in downstream tasks.
	
	\subsection{Explainability}
	Despite that FMs are currently used as interpreters, in many works, to illustrate why anomalies are detected by providing textual explanations in a human-understandable manner, the models themselves are not transparent or interpretable enough to give explanations of the internal model structures and behaviors. However, providing evidence and logical steps for decision-making is a critical issue in anomaly detection especially in applications like healthcare and finance, where understanding the reasoning behind an anomaly is as important as detecting it~\citep{jin2025logicad}. Therefore, how to improve the interpretability of FMs remains an open research question, and it is absolutely important for researchers to enhance the trustworthiness, usability, and decision-making transparency of anomaly detection.

	\subsection{Multimodality}
	While multimodality is considered a critical element of intelligence, and serves as a crucial component for the development of both thorough and broad comprehension of the world, models that go beyond simple alignment of vision and language are yet to emerge~\citep{radford2021learning, an2025gpt}. In real-world application scenarios, taking healthcare as an example, medical data are highly multimodal, with various data types, scales, and styles. Multimodal models generally have better performance on anomaly detection by integrating and analyzing data from different sources than models with a single modality. However, current anomaly detection models are mainly developed for single modality (e.g., text, image, and gene), and do not learn from various modalities. By harnessing information across different data types, multimodal data can provide richer and more robust anomaly detection, thereby improving the accuracy of detection results across various applications. Future studies should further examine the design of FMs that integrate various modalities and domains.
	\section{Conclusion}
	The application of FMs to anomaly detection has emerged as a prominent area of research in recent years.
	In this survey, we provided an in-depth review of the use of FMs in anomaly detection tasks. We introduced a novel taxonomy classifying the methods into three categories based on the roles FMs may play in different stages of anomaly detection, namely encoder, detector, and interpreter. This taxonomy clarifies how FMs can empower the process of data representations, anomaly detection, and explainable analysis of detection results. We also discussed challenges and highlighted several future research directions, aiming to shed light on the way ahead in the field of anomaly detection with FMs. As FMs continue to evolve, their role in anomaly detection is expected to expand, unlocking new possibilities for more accurate, interpretable, and scalable anomaly detection systems.

\vspace*{-12pt}

\bibliographystyle{plain}  
\bibliography{references}

\end{document}